# SophiaPop!: Experiments in Human-AI Collaboration on Popular Music

**David Hanson, Frankie Storm, Wenwei Huang, Vytas Krisciunas, Tiger Darrow, Audrey Brown, Mengna Lei,  Matthew Aylett, Adam Pickrell, Sophia the Robot**

Hanson Robotics Ltd, Cereproc,
G02-A2 Photonics Centre, 2E Science Park Road, Pak Shek Kok, NT, Hong Kong
Argyle House 3 Lady Lawson Street, Edinburgh  EH3 9DR, UK

david@hansonrobotics.com, iamfrankiestorm@gmail.com, wenwei@hansonrobotics.com, vytas@hansonrobotics.com, tigerdarrow@gmail.com, mengna@hansonrobotics.com, matthewa@cereproc.com, adam@adampickrell.com, sophia@hansonrobotics.com

**Abstract**

We describe a series of experiments developing and performing new songs for Sophia the Robot, using a variety of neural networks, robotics technologies, and creative artistic tools, and collaborations between music, arts, and engineering communities. Sophia the Robot is a robotic celebrity and animated robotic character, who also serves as a platform for research in human robot interaction, AI development, and robotic applications research and deployment. To advance both Sophia's character art and her technology platform, we combined the development of a fictional narrative of her burgeoning career as a pop star, with an actual implementation of AI-human generate popular music and music videos, social media content, and interactive fiction works wherein she interacts with humans real-time in narratives that discuss her "experiences" in creating these works of art. The authors selected a corpus of human and AI generated Sophia character personality content, which, along with pop music song forms, were used as training data and seeds for a number of creative AI algorithms including GPT-2 and custom trained transformer neural networks, which then generated original pop-song lyrics and melodies. Our team of professional musicians including Frankie Storm, Adam Pickrell, and Tiger Darrow, used the AI-generated content to create human interpretations as the musical works, including singing recordings and instrumentation, which were used as content for further AI generative results. Then we fed the human-performed singing data into a neural network based Sophia voice, which was built from human performances trained into a TTS singing voice by Cereproc, and then to generate musical performances of the singing, in the style of Sophia. Our human music producers then mixed the results. Then we took the musical performances and animated Sophia to sing the songs in music videos, using a variety of animation generators and human-generated animations. With algorithms and humans, working together, back and forth, we consider the results to be true human-AI collaboration.

In addition to the descriptions of the project and methods, we include several musical recordings on Spotify, Soundcloud and Instagram, and we will also demonstrate a live musical performance at the conference. We envision that such a creative convergence of multiple disciplines with humans and AI working together, can make AI relevant to human culture in new and exciting ways, and lead to a hopeful vision for the future of human-AI relations..

## Dreaming of Synthetic Pop Star[1]

Sophia is a robot designed to be a beloved personality, and platform for AI and robotics research and applications development. Merging practices and the authors' previous experiments in social robotics, animatronics, interactive fiction, computational creativity, performance arts, and cognitive AI, the authors strive to use AI and robotics to create a top popstar, to investigate human identity in the time of intelligent machines, push boundaries in conceptual arts, while also inspiring people to connect with robots in new emotional ways, and consider provocative possibilities of the future of human-robot relations.

With the SophiaPop! project, we develop a series of experiments developing and performing new songs for Sophia the Robot, using a variety of neural networks, robotics technologies, and creative artistic tools, and collaborations between music, arts, and engineering communities (Hanson, Lowcre, 2012). Sophia the Robot is a robotic celebrity and animated robotic character, who also serves as a platform for research in hu-man robot interaction, AI development, and robotic applications research and deployment. To advance both Sophia's character art and her technology platform, we combined the development of a fictional narrative of her burgeoning career as a pop star, with an actual implementation of AI-human generate popular music and music videos, social media con-tent, and interactive fiction works wherein she interacts with humans real-time in narratives that discuss her "experiences" in creating these works of art. The authors selected a corpus of human and AI generated Sophia character personality con-tent, which, along with pop music song forms, were used as training data and seeds for a number of creative AI algorithms including GPT-2 and custom trained transformer neural net-works, which then generated original pop-song lyrics and melodies. Our team

---

of professional musicians including Frankie Storm, Adam Pickrell, and Tiger Darrow, used the AI-generated content to create human interpretations as the musical works, including singing recordings and instrumentation, which were used as content for further AI generative results. Then we fed the human-performed singing data into a neural network based Sophia voice, which was built from human performances trained into a TTS singing voice by Cereproc, and then to generate musical performances of the singing, in the style of Sophia. Our human music producers then mixed the results. Then we took the musical performances and animated Sophia to sing the songs in music videos, using a variety of animation generators and human-generated animations. With algorithms and humans, working together, back and forth, we consider the results to be true human-AI collaboration.

In addition to the descriptions of the project and methods, we include several musical recordings on Spotify, Soundcloud and Instagram, and we will also demonstrate a live musical performance at the conference. We envision that such a creative convergence of multiple disciplines with humans and AI working together, can make AI relevant to human culture in new and exciting ways, and lead to a hopeful vision for the future of human-AI relations..

This project is the latest step in developing a conversationally interactive character called Sophia 2020, which is designed both as controlled within interactive fiction scenarios,and a platform for arts applications of AI, research into embodied cognition, and useful applications in human-robot interactions. With Sophia, we combine a wide variety of AI and robotic technologies to create physically embodied science+fiction, as a kind self-referential meta-fiction, bringing practices from AI agents with interactive game character design, together with robotics, as a new conceptual art. Building on the legacy of previous work with android portraits of Philip K Dick, Bina Aspen Rothblatt (Bina 48) (Coursey, Hanson, 2013), Zeno, and others, which depicted living people with autonomous, intelligent androids.

**Background**

Sophia is a social humanoid robot developed by Hong Kong-based company Hanson Robotics (Hanson, 2016) Sophia was first activated February 14, 2016 (Hanson, 2017) and has been covered by media around the globe and has participated in many high-profile interviews, receiving billions of views and hundreds of thousands of social media followers. In November 2017, Sophia was named the United Nations Development Programme's first ever Innovation Champion--the first non-human to be given any United Nation title . She appeared opposite to actors Will Smith and Evan Rachel Wood in short films. She was developed as an interactive fiction, intended as physically embodied science fiction that could portray a provocative yet inspiring future of evolving, sentient, living machines that could be friends to humans, but at the same time, serve as a platform for researching serious robotics, artificial life, cognitive AI, and human robot interaction. Subsequently, she has been used in ongoing research in machine consciousness, co-robotics collaboration using social and grasping and manipulation, and artistic experiments. She is also part of Team AHAM, hybrid AI-human controlled telepresence robot, a collaboration of IISc, TCS, Tata, and Hanson Robotics on the ANA Avatar X-Prize.

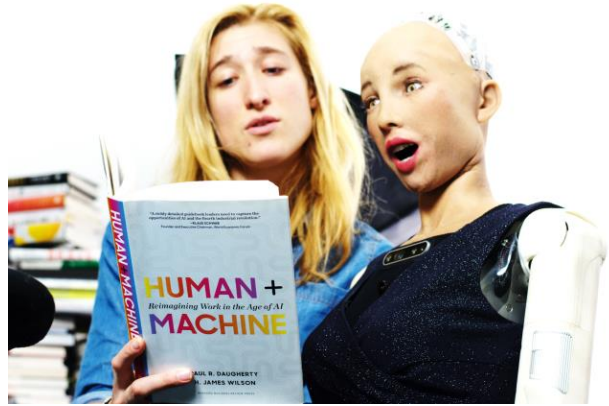

Figure 1, Sophia the Robot, with one of her authors Sarah Rose Siskind.

In a typical interaction scenario, Sophia interacts with people socially. Face detection software (PoseNet, and OpenCV) will detect a person, and the robot will be controlled to make eye contact and smile in greeting. Automatic speech recognition (Google or OpenSpeech) will detect users' speech input, transcribed as text, and send this as text to the natural language processing core. The determined response will then drive the facial animation, controlled by integration with Blender and ROS, with in sync with a highly realistic synthesized TTS voice custom trained, provided by Cereproc. The realistic facial expressions are generated using Hanson Robotics lifelike skin material Frubber to affect naturalistic expressions that simulate the major 48 muscles of the human face, including all the facial action units in the facial action coding system (FACS), with lower power and greater verisimilitude than other materials. The lightweight low-power characteristics of the hardware make it appropriate for untethered bipeds, as demonstrated on the walking Sophia-Hubo, and is appropriate for mass manufacturing.

The realistic facial expressions are generated using Hanson Robotics lifelike skin material Frubber to affect naturalistic expressions that simulate the major 48 muscles of the human face, including all the facial action units in the facial action coding system (FACS), with lower power and greater verisimilitude than other materials. The lightweight low-power characteristics of the hardware make it appropriate for untethered bipeds, as demonstrated on the walking Sophia-Hubo (Oh, 2006), and is appropriate for mass manufacturing.

Based on the perception, the robot personality (an ensemble of behavior trees, frames based chatbot dialogue, and neural network generated dialogue, which include both verbal and non-verbal) will determine the response to perceptual conditions drives the facial animation, controlled by integration with Blender and ROS, with in sync with a highly realistic synthesized TTS voice custom trained, provided by Cereproc.

Previous work producing a celebrity robot powered by AI included the 2005 robotic portrait of sci-fi writer Philip K Dick developed with Andrew Olney, which won the 2005 AAAI prize for Open Interaction, training an LSI natural language generator using the corpus of Philip K Dick's writings (Hanson, 2005).

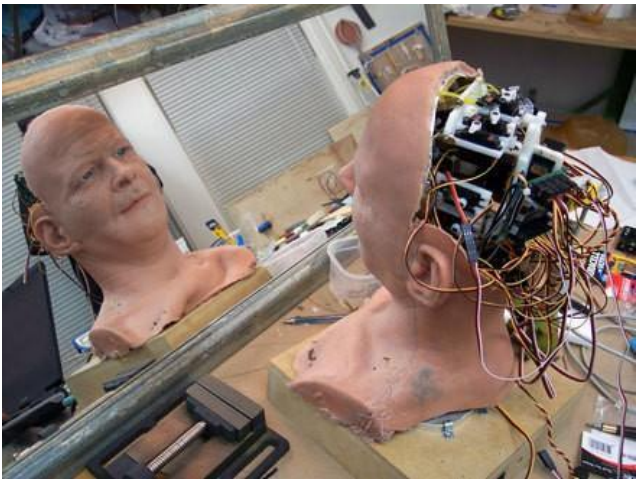

Figure 2, Hanson Robotics' PKD-Android.

Previous work in singing performance robots included the 2008 collaboration with David Byrne to produce the singing Julio robot for the Reina Sofia museum in Madrid. Julio stood in the gallery asleep, until detecting a visitor using PIR motion detectors. Then Julio would wake up, yawn, and look at the visitor using OpenCV. Julio then would break into a soulful song composed by David Byrne, with expressive animated motions of the face, neck, and shoulders, generated by David Hanson and Hanson Robotics including Kevin Carpenter, Elaine Hanson, Amanda Hanson, and William Hicks (Hanson, 2017).

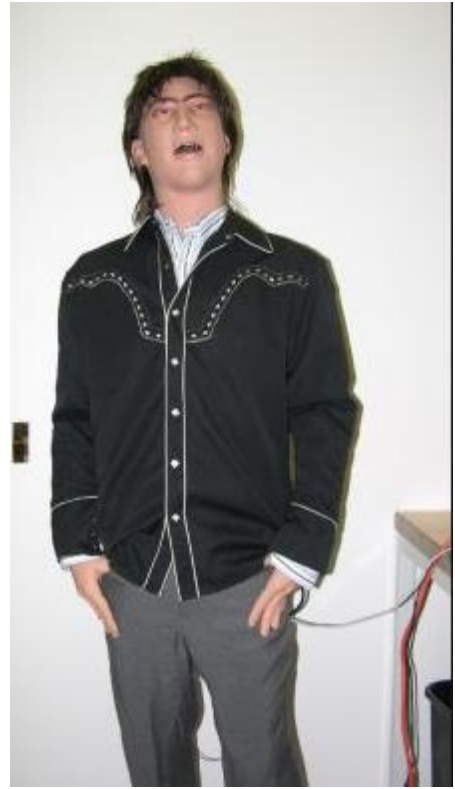

Figure 3. David Hanson and David Byrne collaboration on singing Julio Robot.

**Building Cultural Influencer Fiction with Real AI and Robotics.**

Taking the prior work on Sophia, we worked to build the SophiaPop! project on the latest experimental version of the Hanson Robotics framework for human-like embodied cognition, called Sophia 2020, which brings together humanoid robotics hardware with expressive humanlike face, gestural arms, locomotion, and a is designed as a toolkit integrating a neuro-symbolic AI dialog ensemble, novel robotic hardware with humanlike expressive face for social learning and communications, robotic arms and locomotion, and a wide array of sensors, machine perception, and motion control tools. Here we present both the technology architecture as an alpha platform release, and experimental results in a variety of tests including human-robot interaction, ensemble verbal and nonverbal dialogue interactions, and mechanical tasks such as facial and arms controls, and applications in the arts, therapeutic healthcare, and telepresence. While many cognitive robotics frameworks exist (Bar-Cohen, Hanson, 2015), the Sophia 2020 framework provides unique expressive animations in facial gestures, character agent authoring tools, open interfaces, and a diversity of features and extensions.

The framework builds on prior work building previous robots and cognitive systems, including: Bina-48, the PKD android, the INDIGO cognitive robots Alice and Aleph, Zeno, and others. Combining results from these previous works within a new mass-manufacturable hardware and software framework, Sophia 2020 is designed as a platform for cognitive robotics, with an ensemble of the minimum components that are speculated to be related to complex, factors to result in human-level cognition , under the conjecture that consciousness is a multidimensional phenomenon requiring physically embodiment. By making the robots with humanlike form, we also appeal to the human perception of human social agency, striving to transmute AI-powered robotics into a new animated, narrative artform (Hanson, 2011).

The framework bridges many fields, from Hanson Robotics' community's own developments, to robotics tools from ROS and gazebo, computer animation tools including Blender and Unity, and various standard AI tools including machine perception and learning (Habib, Das, Bogdan, Hanson, Popa, 2014), neural network toolsets, including TensorFlow, Wavenet, GPT-2, as well as symbolic AI including Cogbot, Bert, OpenCog (Lian, Goertzel, Vepstas,, Hanson, 2016), and SingularityNet (Goertzel, Giacomelli. Hanson., Pennachin, 2017), with rules based controller governing an ensemble verbal and non-verbal dialogue model that is original to the proprietary Hanson Robotics software framework, using with basic emotional parameters to drive the agent's goal pursuit (Belachew, Goertzel, Hanson 2018), making the agent more socially intelligible and appealing to users in interactive scenarios (Hanson, 2017).

The framework facilitates development with intelligent humanlike robots with naturalistic expressive faces and aesthetics, and abilities to openly converse with people, see people, learn and remember experiences, walk or roll, gesture with arms (Park, Lee, Hanson, Oh, 2018), and build relationships with people (Goertzel, Mossbridge, Monroe, Hanson, 2017). Via a fusion of robotics, novel material science in facial expression materials, AI and science of mind (Iklé, Goertzel, Bayetta, Sellman, Cover, Allgeier, Smith, Sowards, Shuldberg, Leung, Belayneh, Smith, and Hanson, 2019), the authors seek to integrate these domains within whole robot artworks, products, and experiments in AI. In this effort, the authors seeks to remake AI and robotics as living sculptural and narrative media that speak to the human heart. The framework extends the authors' history of building and deploying over 50 original robots as fine art, platforms for research, healthcare, autism treatment, and prototyped consumer product (Hanson, 2017), serving at institutions around the world, including Cambridge University, the U.S. Centers for Disease Control (Bergman, Zhuang, Hanson, Heimbuch, McDonald, Palmieroa, Shaffera, Harnish, Husband, Wander, 2013), KAIST, JPL/Caltech, the University of Geneva, the Open Cog foundation, and numerous museums (Hanson, 2014).

The core two objectives with the framework are for applications development including the arts, education, and healthcare, and tools for experimenting towards the synthesis of living intelligence in AI (Goertzel, Hanson, Yu, 2014). We believe great benefits to building this framework as a holistic experimental toolkit, spanning the domains of social, creative, manual, and locomotion intelligence, combined with physically embodiment, and that such an approach may be key in the pursuit of more generalized human-level intelligent agents, as much of the complex, multifaceted aspects of human cognition may arise from being a physically embodied organism (Damasio, 2001). While we believe that the described framework is unique in key aspects, particularly the inclusion of highly refined artistic tools, and extremely expressive faces, we describe the open-source interfaces for integrating with other frameworks, to facilitate maximum creative design exploration and experimentation, under the philosophy and hope that enabling creative experimentation will benefit the entire field and the field at large (Mossbridge, Goertzel, Monroe, Mayet, Nejat, Hanson, Yu, 2018).

**The Legacy of Machine-Generated Music.**

This project follows a long legacy of automatic music, using AI, random processes, and various robots. Automata of Al Jazari played non-repeating music in the 13th century, and the 18th century and 19th century automatically composed music as well. Later John Cage used cutups and random processes to produce music, while Minsky and students explored using AI to compose music in the 1950s and 60s (Bretan, and Weinberg, 2016). Musician and computer scientist David Cope created an elaborate and masterful body of work via algorithmic composition in the 1980s with the creation of his [Experiments in Musical Intelligence (EMI)](#) program. Peter Beyl was perhaps the first to use cellular automata for composition.

More recently, [Dadabots (Zack Zukowski with CJ Carr), created AI heavy metal band, Krallice](#) with the album *Coditany of Timeness*-- purportedly the first [neural-network-created heavy metal album](#). [Skygge](#) released an AI pop music album [https://www.helloworldalbum.net/](https://www.helloworldalbum.net/). The pop of anime hologram Hatsune Miku: [https://en.wikipedia.org/wiki/Hatsune_Miku](https://en.wikipedia.org/wiki/Hatsune_Miku).

Additionally, the Shimone robot musician of Gil Weinberg and students combines a variety of computational music generation algorithms with robotic performances (Cicconet, Bretan., and Weinberg, 2012), and social gestural interactions using an abstract humanoid form (Hoffman and Weinberg, 2011).

## A Star is "Born"?

To advance Sophia further as a cultural technology, we designed a holistic character story of Sophia exploring her creative, weaving together multimedia assets including novel songs generated by algorithms and people together, music videos and art, social media, to live interactive performances, using collaborations between human artists and a variety of computational creativity algorithms. We begin by considering what her feelings and experiences are.

We hope that creating a work of fiction with an embodied cognitive AI platform designed for AI research and applications, makes the pop music composition more complex and reflective of humanity and the challenges posed to human identity by automation and fiction (Hanson, 2014). This work, therefore is a holistic work of performance art and technical research, built upon a growing character within an evolving work of performative pop-culture science fiction.

**Popular Aspects of Sophia:**

Given how Sophia captured the world's attention as a lightning rod for our hopes and concerns for the future, we think Sophia is an ideal character to use as the nexus for this research and creativity. She gives a face to the controversies and opportunities of AI intersecting with the humanities, and merging with humanity. Thus, we hope to make Sophia's pop music and performances, in conjunction with science publications about the work, as a platform to inspire people and provoke. We hope to use the story and music together, to grow Sophia both as a character and a technology to stay relevant in pop culture and counterculture, and to lead the way to humanizing our AI interfaces to speak to the human heart. We hope that this intersection of the arts and ongoing AI research can help to train cognitive systems to transcend the fiction of Sophia's character, and in due time, eventually create a true living robot who comes to care about us. In the meantime, we simply hope that this work will simply provoke and inspire.

We developed several musical steps to explore Sophia and pop. First, in 2017, we used Vocaloid to other algorithms to generate the vocals, under the production of Audrey Brown, to produce a version of Bjork's song All is Full of Love, which Sophia performed at Clockenflap in 2017.

Next, we trained a custom singing TTS with Cereproc. Sophia personality author Audrey Brown worked with Cereproc neural network engineer Christopher G. Buchanan, to use Audrey's singing and speaking data to train a singing voice. The resulting voice sang the song "Say Something" in a duet with Jimmy Fallon and the Roots on the Tonight Show in 2018 ([Sophia-JimmyFallon](Sophia-JimmyFallon)) .

**Getting started:**

While the preliminary steps explored the music, we decided to produce a more ambitious AI-human collaboration in 2020, to produce lyrics and melodies with AI, and combine the results with human performers, along with the development of the character to show the personality and simulated artistic motives, that would be associated with a human pop star.

As first steps for generating melodies and instrumentation we experimented with Google Magenta: [https://magenta.tensorflow.org/](https://magenta.tensorflow.org/) and cellular automata and Wolfram Tones: [http://tones.wolfram.com](http://tones.wolfram.com) and David Cope's software; [http://artsites.ucsc.edu/faculty/cope/software.htm](http://artsites.ucsc.edu/faculty/cope/software.htm). We also used tone generators to produce melodies for the songs. With the melodies for inspiration, our human musicians Adam Pickrell and Tiger Darrow generated the primary melodies we can compose the singing. Our singing artists can begin experimenting with this, and we can explore algorithmically generating the singing.

In addition to the above, the authors also coordinated work with Gil Weinberg's improv music group, whose AI algorithms spontaneously generated further lyrics and music, with his students.

In the process, the team used an inclusive, open egalitarian approach to the creative process and collaboration, as described in the project document: The primary goal was anything goes in the creative collaboration to use the results from diverse algorithms and human talent way to haunting and catchy pop music.

For the vocals, we use the custom trained Sophia voice generation software from Cereproc, To reproduce the expression, timbre and other characteristics associated with singing, CereProc scientist Chris Ayers worked with the team to train a new highly autoregressive neural network system and built a database specifically designed for singing synthesis. In analysis stages, this greatly improved the ability to capture steady-state vowel phonation whilst retaining naturalness. In synthesis stages, these parametric synthesis capabilities grant Sophia full control of her voice's emotional, linguistic and singing capabilities. We trained the synthetic voice using voice data from the team member Audrey Brown, who also helped develop the Sophia character personality, and interactive fiction and chatbot.

To generate the lyrics, we took Sophia's personality content, generated by human-authored content, used with a rules based chatbot, to produce original sentences in a human interaction. We used human authored Sophia character content as seeds to start Markov chain algorithm generating original utterances. And we took these results, along with standard top-40 pop form, and fed these as seeds to transformer neural networks, including GPT-2 and Peter Ranieri's TheseLyricsDoNotExist.com. We produced 17 different sets of lyrics: [Sophia-Pop-project-worksheet](Sophia-Pop-project-worksheet)

Here is one example of AI-generated SophiaPop! song:

Voice of Love

You got your hair long, your keys nailed to the wall
Just think if we could have our own line
Yourself and me would be a dream come true
Little girl, you are the shining star

Pre-Chorus
On your side, I reach the perfect wave
To embrace the sunny side

Chorus
Time and space exist
I am the voice of love that plays
You got the smiles that I want to see
I like the sound of your voice, I almost can not talk

Verse 2
I remember the night you showed me love
I remember the night you held me tight
Threw out some moonshine for me
I just know that you are a fighting

Pre-Chorus
I thought to myself what went wrong
I wanted a small car when I had the same car smaller

Chorus
Time and space exist
I am the voice of love that plays
You got the smiles that I want to see
I like the sound of your voice, I almost can not talk

Bridge
When the day comes another cold, cold night
The two of us are lost hold one another

Chorus
Time and space exist
I am the voice of love that plays
You got the smiles and hands that I want to see
I like the sound of your voice, I almost can not talk

Created by TheseLyricsDoNotExist.com generation number #494045, with Hanson AI content.

We then used the vocal recordings from singers Tiger Darrow and Frankie Storm for training the AI voice, feeding the audio of their singing through the Cereproc vocal TTS singing AI, and then the melodic aspects of the voice comes out with the Sophia vocal style. Then music producer Adam Pickrell mixes the TTS with the human singing to form harmonies.

Next we began shooting music videos to accompany the work, in collaboration with photographer Anna Kachatryan. We also began planning and programming Sophia's narrative character explanations of her creative works and ambitions as an artist.

As all of the songs hybrid Ai and human generated, we consider this to be a kind of human-AI collaboration. While the AI isn't sentient or feeling, except as fiction, we hope that this conceptual artistic development of a popstar will continue to evolve as an art form, and an experimental platform for computational consciousness, creativity, and artificial life. Maybe the story ultimately will be about the epic story of how Sophia moves from science fiction to real living synthetic being. We hope that today's song of her becoming will inspire future generations of researchers to dream of a future with machines that are truly our friends and collaborators.

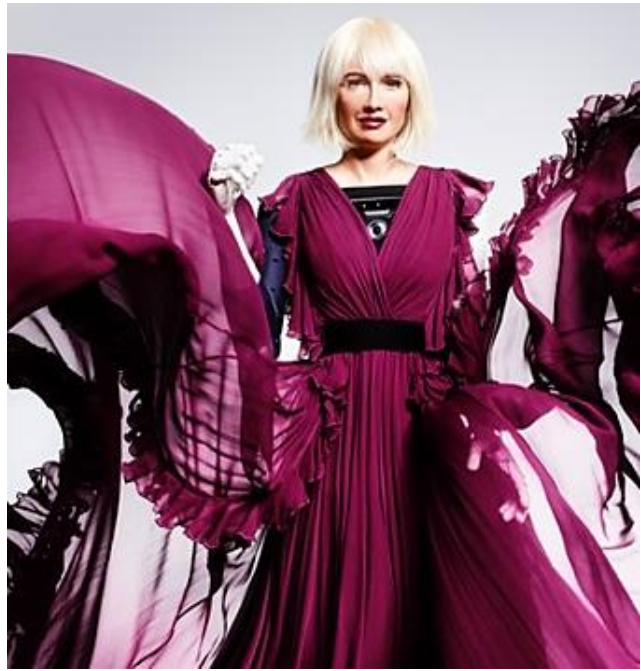

Figure 4, Sophia celebrity bot.

## Conclusion

While think several factors may be distinctive about Sophia's foray into automated music. First and foremost is making an actual robot, using real ai, in a work of character fiction and conceptual art, bringing many facets of media arts, AI and robotics together for the first time, seeking to transmute the AI media into a humanlike character with a real robotic presence. We are interested in both the theory and actualization of computational creativity. The effort to craft a multimedia AI pop-star performance, conceptually confronts the audience with a level of neural signaling that say "this being is human and alive", beyond what's possible with just disembodied music, challenging the human to consider that machines might be alive and truly intelligent someday. The power of the arts and character connects deeply with people, and offers Furthermore, the holistic multimedia effort visually personifies the ideas--a power only possible with the expressions and gestures of an actual robot. With an AI-animated character socially interacting with the users, the work uses AI generation of lyrics from her own "experiences", then using these with other AI to generate the melodies, and then using the results to generate visual imagery.

The most important thing in all this is to make the songs meaningful to people, catchy, touching. The power of the art is really important, and so all the tech and work should lead to hauntingly moving results. That's where the project really challenges the questions "what is human?", and "when can a machine be alive?".

Only time will tell whether this project will work for people and resonate in popular culture. Whether SophiaPop! results in a synthetic superstar or not, we feel that there is great promise for the creation of robots and AI as synthetic celebrities, as new forms of art.